\documentclass[letterpaper, 10 pt, conference]{ieeeconf}  

\IEEEoverridecommandlockouts                              

\usepackage{cite}  
\usepackage{amsmath,amssymb,amsfonts}
\usepackage{graphicx}
\usepackage{textcomp}
\usepackage{xcolor}
\usepackage{url}
\usepackage[ruled,vlined]{algorithm2e}
\makeatletter
\let\NAT@parse\undefined
\makeatother
\usepackage[colorlinks=true,
linkcolor=blue,
citecolor=blue,
urlcolor=blue,
pdfborder={0 0 1} 
]{hyperref}
\usepackage{cleveref}
\usepackage{threeparttable}
\title{\LARGE \bf
ASC-SW: A Lightweight Atrous Strip Convolution Network for DLOs Segmentation on Edge mobile Robots
}

\author{}

\author{Cheng Liu$^{1}$, Fan Zhu$^{1}$, Yifeng Xu$^{2}$, Baoru Huang$^{3}$, Mohd Rizal Arshad$^{1}$
\thanks{$^{1}$Cheng Liu, Fan Zhu and Mohd Rizal Arshad is with School of Robotics,
        Xi’an Jiaotong-Liverpool University, 215000 Suzhou, China
        {\tt\small Cheng.Liu23@student.xjtlu.edu.cn; Fan.Zhu@xjtlu.edu.cn; Rizal.Arshad@xjtlu.edu.cn}}%
\thanks{$^{2}$Yifeng Xu is with the Department of Robotics, University of Michigan, Ann Arbor, USA
        {\tt\small yifengxu@umich.edu}}%
\thanks{$^{3}$Baoru Huang is with the Department of Artificial Intelligence, University of Liverpool,
        Liverpool, UK
        {\tt\small Baoru.Huang@liverpool.ac.uk}}%
}

\begin{document}

\maketitle
\thispagestyle{empty}
\pagestyle{empty}
\begin{abstract}
Detecting deformable linear objects (DLOs), such as floor cables, is essential for safe mobile robot navigation but remains challenging due to oblique viewpoints, thin structures, and limited edge-device resources. Existing DLO segmentation methods are primarily designed for manipulator platforms with fixed top-down views and often require heavy models, limiting their deployment on mobile robots.
We formulate a cross-view DLO segmentation problem, where models trained on manipulator-view data must generalize to mobile robot perspectives. To address this, we propose ASC-SW, a lightweight and geometry-aware segmentation framework. The core network, ASCNet, introduces Atrous Strip Convolution, combining directional strip filtering with dilated receptive fields to enhance sensitivity to elongated structures at low computational cost. An Atrous Strip Convolution Spatial Pyramid Pooling module enables multi-scale anisotropic feature aggregation, while a temporal Sliding Window refinement suppresses viewpoint-induced false positives.
Evaluated on real-world mobile robot data, ASC-SW achieves 74.1\% mIoU at 261 FPS and remains deployable on edge devices.

\end{abstract}
\section{Introduction}
Recent advances in mobile robotics, including SLAM \cite{dhaigude2025comprehensive} and autonomous navigation, have been substantial. However, conventional LiDAR-based systems \cite{vinci2025lidar}\cite{chen2025review} still struggle to detect near-ground obstacles, such as cables on the ground or surface depressions.
Although extensive research on manipulator platforms has addressed the perception and handling of deformable linear objects (DLOs)\cite{han2026robotic}\cite{vodolazskii2025review}, analogous investigations within the domain of mobile robots remain relatively underexplored. On manipulator platforms, researchers have conducted extensive studies on DLOs. Previous research \cite{caporali2022fastdlo}\cite{caporali2023deformable}\cite{caporali2024deformable}\cite{holevsovsky2024movingcables}\cite{lau2024large}\cite{caporali2025robotic}\cite{caporali2024dlo}\cite{xiang2023trackdlo} has focused on the real-time segmentation of instances and tracking of visible objects from a fixed viewpoint, enabling the manipulator to interact with these objects. In related work on DLOs, many approaches begin by using segmentation models to perform an initial segmentation of the object. Some studies employed the CNN model for segmentation\cite{caporali2022ariadne+}\cite{caporali2023rt}\cite{caporali2022fastdlo}\cite{caporali2025robotic}. Other works\cite{zhaole2023robust}\cite{Kozlovsky2024ISCUTEIS} leveraged the Segment Anything (SAM) \cite{kirillov2023segany}. However, both approaches share a common limitation: they cannot be deployed on edge devices due to their high computational cost. 
\begin{figure}[!htb]
            \centering
           \includegraphics[width=\columnwidth]{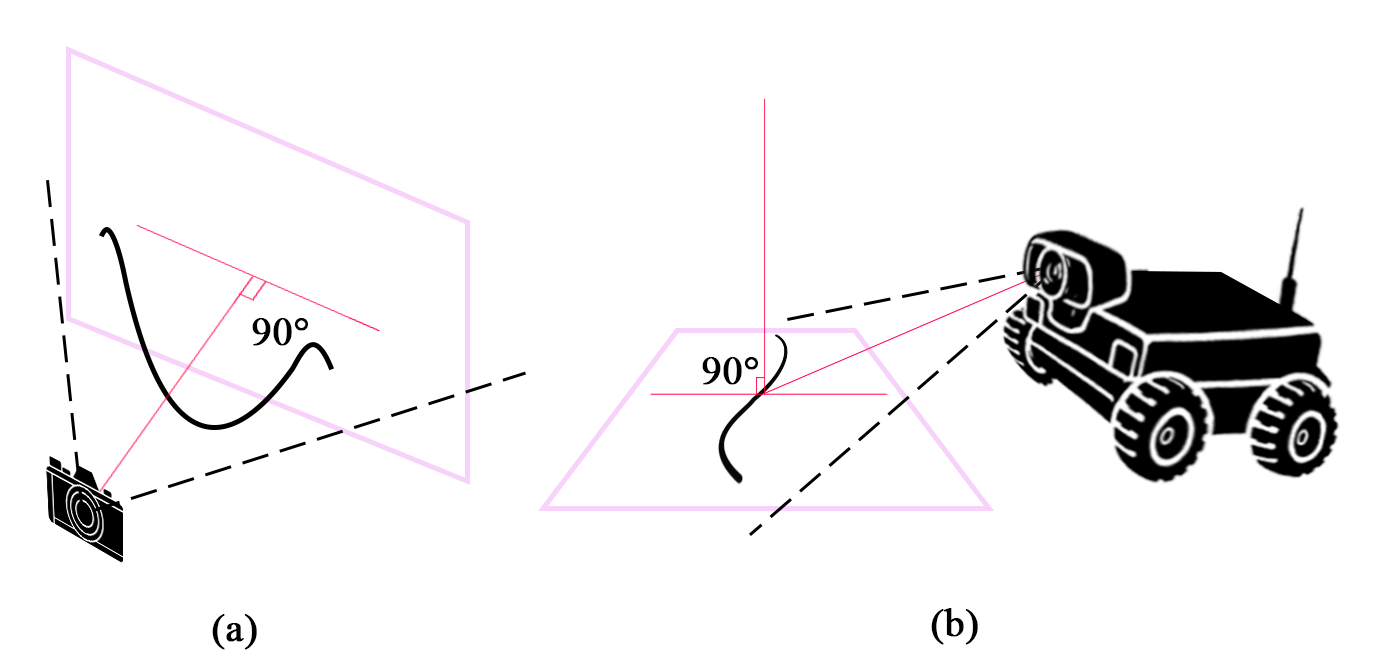}  
            \caption{(a) The DLO detection on the manipulator platform relies on camera views perpendicular to the DLOs plane.
            (b) The mobile robot platform observes DLOs from oblique views.
            Due to this viewpoint shift, models trained on manipulator data perform poorly on mobile robot images.
            Our method enables accurate detection under the mobile robot viewpoint using manipulator data \cite{zanella2021auto}\cite{caporali2023weakly}.}
            \label{fig:perspective}
\end{figure}
Moreover, whether it is SAM or self-trained visual segmentation neural networks, the data they used\cite{zanella2021auto}\cite{caporali2023weakly} are often captured from a viewpoint fixed perpendicular to the plane where the DLOs is located, which is difficult to apply in mobile robot navigation scenarios, and their methods tend to focus on scenes with simple and low-noise backgrounds. If these methods are applied to real-world mobile robots, the segmentation performance degrades due to varying viewpoints and the presence of many distractors and false positives in complex environments. Because the camera view of the mobile robot is different from that of the manipulator platform as shown in Fig.\ref{fig:perspective}, and the manipulator platform's operation does not take into account the situation when the camera position moves.          
\begin{figure*}[!htb]
            \centering
            \includegraphics[width=\textwidth]{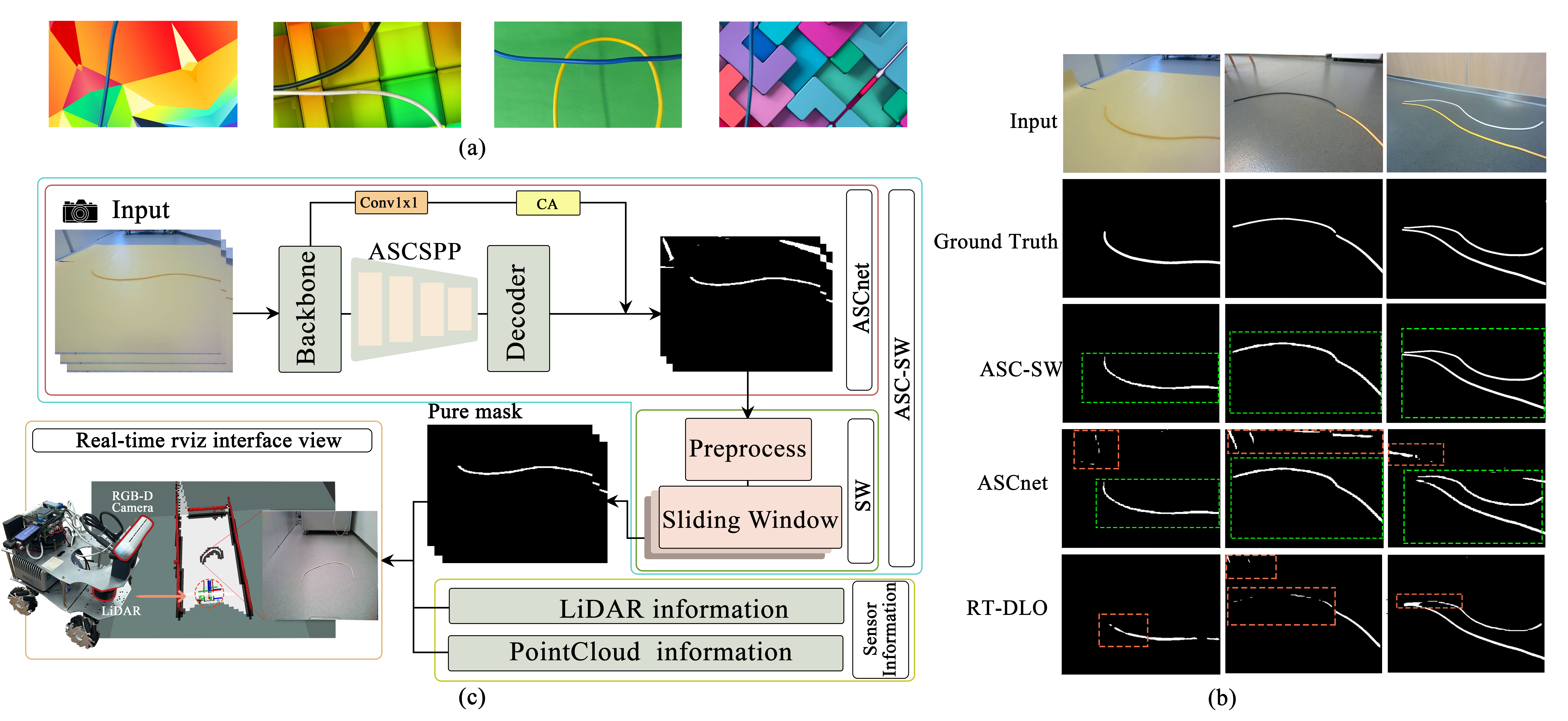}  
            \caption{(a): The synthetic dataset to simulate the viewpoint from the fixed camera on the manipulator (see Fig. 2a). The dataset is utilized as the training data in our method
(b): The results of DLOs detections from the viewpoint (see Fig.\ref{fig:perspective}b) of the mobile robot generated by different methods. Compared with RT-DLO\cite{caporali2023rt}, ASCNet achieves the better segmentation under the low color contrast.The proposed ASCNet produces the initial raw predictions, which are subsequently refined by the ASC-SW sliding-window post-processing module. This process effectively suppresses noise arising from the mistaken segmentation of background objects (red boxes).
(c): The pipeline of our method RGB images are first segmented by the deep neural network ASCNet. The mask output by the neural network is first preprocessed using a morphological erosion operation to refine the segmentation, and is then further optimized by the SW module. The optimized mask is used to filter and downsample the depth point cloud, which is then incorporated into the costmap \cite{lu2014layered} as obstacles for collision avoidance.}
            \label{fig:pipline}
        \end{figure*}
To address the limitations of current mobile robots in avoiding obstacles on the ground, this paper proposes an efficient segmentation model called Atrous Strip-Convolution Sliding Window (ASC-SW) as shown in Fig.\ref{fig:pipline}, which consists of two components: Atrous Strip Convolution Network (ASCNet) and a post-processing algorithm named a Sliding Window (SW) module. We propose the ASCNet, inspired by the DeepLabV3+ model \cite{chen2018encoder}. By decomposing standard convolutions into strip convolutions and combining them with atrous convolutions, we introduce the Atrous Strip Convolution and Atrous Strip Convolution Spatial Pyramid Pooling (ASCSPP). We trained our model without mobile robot viewpoint images and the model for detection of DLOs outperforms the existing DLOs segmentation models on the self-built real world dataset and has been deployed on a physical mobile robot, and enables mobile robots to effectively avoid the DLOs on the ground. 

Our contributions consist of the following parts: 

\begin{itemize}
\item We formulate a novel cross-view DLO detection task, addressing the viewpoint discrepancy between manipulator platforms and mobile robots.
\item We propose ASCSPP, a geometry-aware multi-scale feature extraction module that integrates atrous and strip convolution to enhance directional sensitivity while preserving lightweight efficiency.
\item We introduce a lightweight temporal Sliding Window (SW) refinement module that improves robustness under noisy mobile robot environments.
\item We validate the system on real mobile robot hardware, demonstrating cross-view generalization and edge-device deployability.
\end{itemize}
\section{related work}
\begin{figure*}[!htb]
            \centering
            \includegraphics[width=\textwidth]{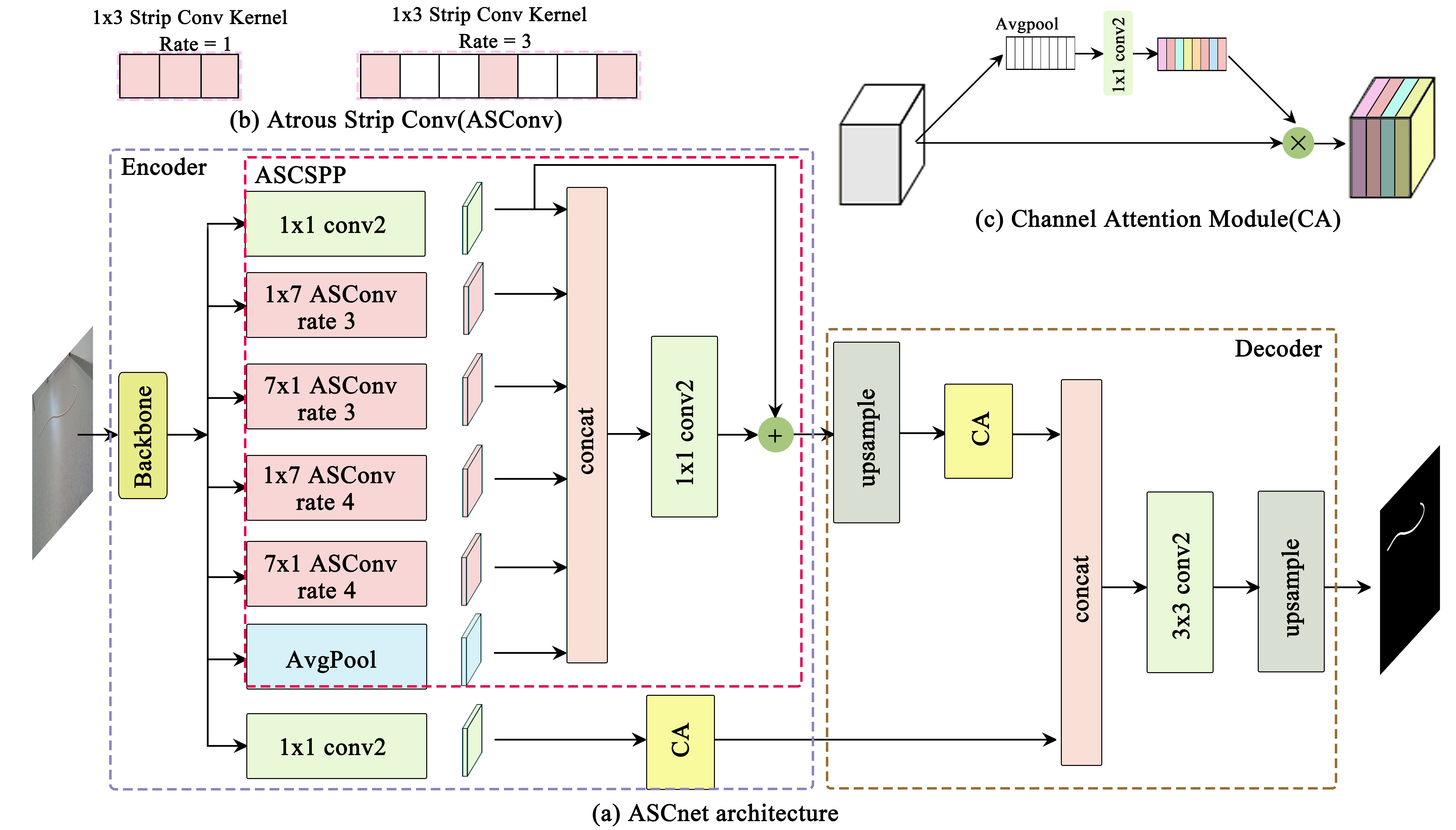}  
            \caption{(a): The structure of ASCNet consists of an encoder and a decoder. The encoder includes a backbone and an ASCSPP module, where the backbone extracts low-level features and the ASCSPP further extracts high-level features. The extracted features are then passed through the channel attention module shown in (c) before being decoded. (b): The convolution kernel of the atrous strip convolution is shown in the figure. Dilation is incorporated into the strip convolution to enlarge the receptive field while reducing computational complexity. (c): The channel attention module, as illustrated, applies a 1×1 convolution to the pooled channel vectors to generate attention weights, which are then applied to the original feature map via weighted multiplication.}
            \label{fig:neral}
        \end{figure*}
\subsection{Deformable Linear Object Detection}
Holevsovsky et al. \cite{holevsovsky2024movingcables} employed an optical flow–based model with a magnitude threshold to track and predict moving lines, and compiled a dataset of dynamically moving linear objects. Zhaole et al.\cite{zhaole2023robust} introduced a 3D DLO reconstruction method under occlusion, combining SAM-based 2D segmentation with K-means clustering\cite{ahmad2007k} and B-spline\cite{gordon1974b} interpolation to generate 3D DLO chains. Kozlovsky et al. \cite{Kozlovsky2024ISCUTEIS} proposed a framework based on the foundation model for instance segmentation of DLOs , introducing text embeddings via pretrained models such as CLIP enabling real-time keypoint estimation on DLO. Caporali et al. \cite{caporali2024dlo} adopted a foundation-model-based approach for DLO instance segmentation. Xiang et al. \cite{xiang2023trackdlo} used depth information to track DLOs. Caporali et al. 
\cite{caporali2022ariadne+,caporali2023rt,caporali2022fastdlo,caporali2023deformable,caporali2024deformable} employed deep neural networks for binary DLO classification, using post-processing and geometric refinements to achieve real-time instance segmentation by analyzing mask features such as curvature and intersections. In \cite{caporali2025robotic}, They use 2D segmentation to support 3D DLOs tracking. While manipulator platforms rely on 2D DLO segmentation for 3D reconstruction and tracking, our approach adapts existing manipulator data for efficient DLO detection and segmentation from a mobile-robot viewpoint.
\subsection{Strip Convolution and Atrous Convolution}
Guo et al. \cite{guo2023visual} found that splitting large convolutional kernels can still achieve excellent performance. Lau et al.\cite{lau2024large} converted 2D depthwise and depthwise-separable convolution kernels into 1D strip kernels along vertical and horizontal directions. Liao et al.\cite{liao2024sc} designed a neural network with strip convolutions and a strip attention fusion module for semantic segmentation of seedlings and green crops. Guo et al.\cite{guo2022segnext} proposed a semantic segmentation model that replaces traditional convolutions in the attention module with horizontal and vertical strip convolutions and employs a lighter decoder.
Yu et al.\cite{yu2015multi} first proposed atrous (dilated) convolution, using a dilation rate to enlarge the receptive field via sparse input sampling. Chen et al.\cite{chen2017rethinking,chen2018encoder} combined atrous convolution with depthwise separable convolution and the Pyramid Pooling module, and proposed the Atrous Spatial Pyramid Pooling (ASPP) module. Yu et al.\cite{yu2017dilated} introduced the atrous convolution into the ResNet backbone network, expanding the receptive field without reducing spatial resolution. This paper proposes atrous strip convolution, combining atrous and strip convolutions to enhance the CNN’s perception of DLOs while keeping the model lightweight.
\section{method}
\subsection{ASCNet}
The structure of ASCNet is shown in Fig.\ref{fig:neral}a, and it is based on the DeepLabV3+\cite{chen2018encoder} architecture, consisting of a backbone, encoder, decoder, pyramid fusion module, and channel attention module (CA). The backbone is implemented using pretrained MobileNetV2\cite{sandler2018MobileNetV2}. 
We employ the channel attention mechanism \cite{hu2018squeeze} to weight feature maps along the channel dimension. In our design, the fully connected and ReLU layers are replaced with a 1×1 convolution, which preserves the capacity to extract channel-wise attention while substantially reducing computational complexity, thereby meeting the lightweight requirements of the model.
\subsection{Atrous Strip Convolution}
\begin{figure*}[!htb]
            \centering
            \includegraphics[width=\textwidth]{spp.png}  
            \caption{(a): The structure of the ASPP module shown in\cite{chen2018encoder}, (b): The structure of the ASCSPP module proposed in this paper, (c): The structure of the SPASPP module shown in\cite{guo2024dsnet},(d): The structure of the DAPP module shown in\cite{pan2022deep}.}
            \label{fig:spp}
        \end{figure*}
\label{sec:asconv}
ASConv is illustrated in Fig.\ref{fig:neral}b. To better extract features of DLOs, we adopt strip convolutions and integrate them with atrous convolutions to ensure compliance with lightweight model constraints. Building upon the work of Guo et al. \cite{guo2022segnext}, which demonstrated the efficiency of cascaded vertical and horizontal strip convolutions as substitutes for square kernels, we propose a parallel configuration of strip convolutions with varying dilation rates. By decomposing convolutions and varying dilation rates, this method fuses the outputs to extract higher-level linear features.
\subsection{Atrous Strip Convolution Spatial Pyramid Pooling}
Several studies have improved and modified the classic ASPP module\cite{chen2018encoder}. Their structures, along with the ASCSPP proposed in this paper, are shown in Fig.\ref{fig:spp}. It can be seen that SPASPP\cite{guo2024dsnet} and DAPP\cite{pan2022deep} increase the number of convolution operations and apply them in series, and finally add the original input to the concatenated output. These serial operations and repeated stacking of feature maps followed by convolution enrich the extracted features, but also significantly increase the number of parameters and computational cost. We modify ASPP by introducing parallel atrous strip convolutions, which decompose traditional convolutions to better capture fine-grained features.
\subsection{Sliding Window}
        \begin{algorithm}[ht]
        \small 
        \caption{Mask Post-processing with Sliding Window Voting}
        \label{ALG:1}
        \KwIn{Binary mask $M$; Dilation kernel size $(m,n)$; Sliding window size $k$}
        \KwOut{Post-processed mask $M'$}
        $M_\text{erode} \leftarrow \textsc{Erode}\bigl(M,\text{Kernel}(m,n)\bigr)$\;
        $\mathcal{C} \leftarrow \textsc{FindContours}(M_\text{erode})$\;
        $Frame[ID_1,...,ID_n] \leftarrow AssignID(C)$\;
        $Sliding\ window.queue \leftarrow \textsc{Counting}(Frame_k[ID_1...ID_n])$\;
        $[ID_n] \leftarrow \textsc{FindID}(Sliding\ window.queue,K)$\;
         $M' \leftarrow \textsc{Filter}(M_\text{erode},[ID_n])$\; 
        \Return $M'$\;
        \end{algorithm}
Since ASCNet is a lightweight model trained with 224×224 input, it shares the common characteristic of low-resolution trained lightweight models poor resistance to noise interference. These incorrect segmentations act as noise that confuses the robot’s vision and lowers the precision of ASCNet. Therefore, we apply a post‑processing method to the network output to suppress noise and improve detection performance.
        \begin{algorithm}[ht]
        \small 
        \caption{Assign ID}
        \label{ALG:2}
        \KwIn{Counters $[C_1,..,C_n]$; }
        \KwOut{IDs in current $Frame_t$ $[ID_1,..,ID_n]]$}
        $[Cen_1,.,Cen_n] \leftarrow \textsc{Centroid}\bigl([C_1,.,C_n])$\;
         $distance[d_1,.,d_n] \leftarrow \textsc{Euclidean}(Frame_t[Cen_1,.,Cen_n],Frame_\text{t-1}[Cen_1,..,Cen_n])$\;
         \If{$d_n < threshold$}{
            $Tracked\ ID \leftarrow ID_n$\;
          }
          \Else{ $New\ ID \leftarrow ID_n$\; }
         $Frame_t.append(Tracked\ ID,New\ ID)$
        \Return $Frame_t$ $[ID_1,..,ID_n]]$\;
        \end{algorithm}
The algorithm \ref{ALG:1} shows the pseudocode of the entire SW pipeline. The SW takes the mask output by the neural network and first erodes the original mask with a dilation kernel to repair fragmented mask regions. Next, each connected component in the mask is tracked and assigned an ID; The algorithm \ref{ALG:2} provides pseudocodes for tracking and assigning IDs. Each connected component is given an ID and its centroid is calculated. Centroids from the previous frame are compared with those from the current frame and matched using Euclidean distance, components that do not match are assigned new IDs. Then a sliding‑window count of ID occurrences is tallied across frames. Regions with occurrence frequencies in the range [n-k,n], where n is the maximum frequency in the sliding window, are retained, while all others are suppressed to produce the final post-processed mask. From a mobile robot perspective, objects similar to DLOs, such as baseboards, often appear.
 \begin{figure}[htbp]
            \centering
        \includegraphics[width=\columnwidth]{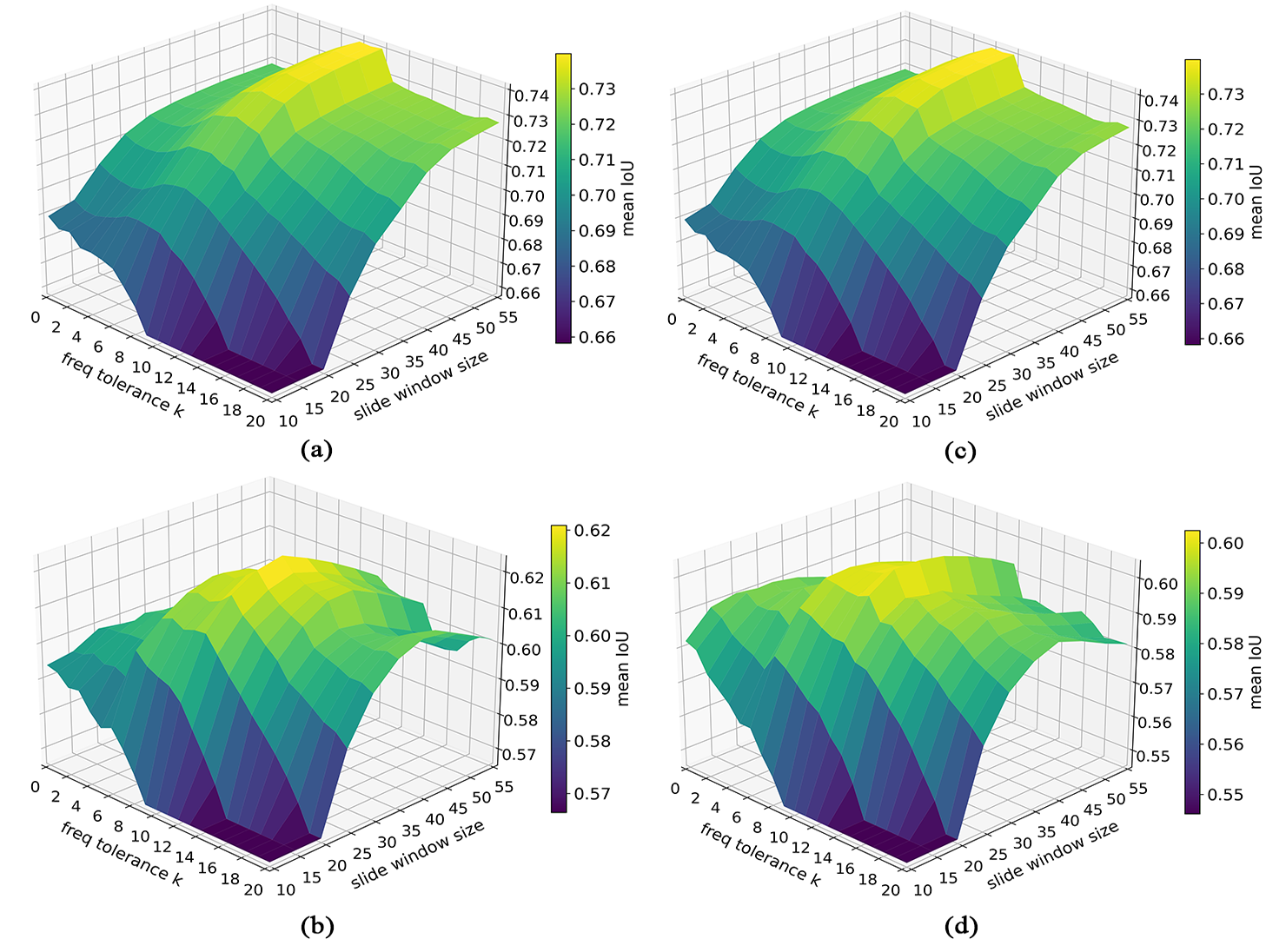} 
            \caption{(a–d): The impact of sliding window size and frequency tolerance K on segmentation performance (kernel sizes: (1,1), (1,2), (2,1), (2,2)).} 
            \label{fig:para}
        \end{figure}
         \begin{figure}[htbp]
            \centering
            \includegraphics[width=\columnwidth]{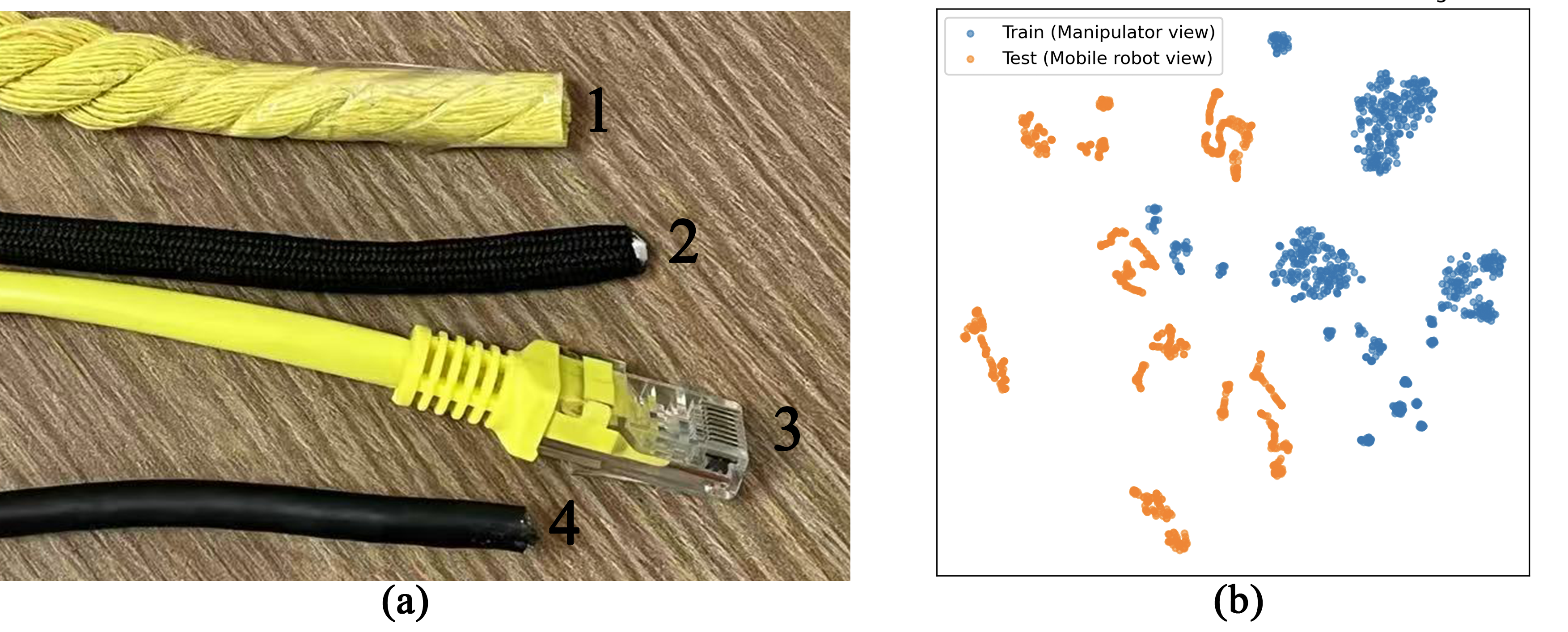} 
            \caption{a: 1: Cotton yarn, 2: polyester rope, 3: ethernet cable 4: data cable. b: The feature distribution between training and testing domains.} 
            \label{fig:lineappearance}
        \end{figure}
\begin{figure*}[!htb]
            \centering
            \includegraphics[width=\textwidth]{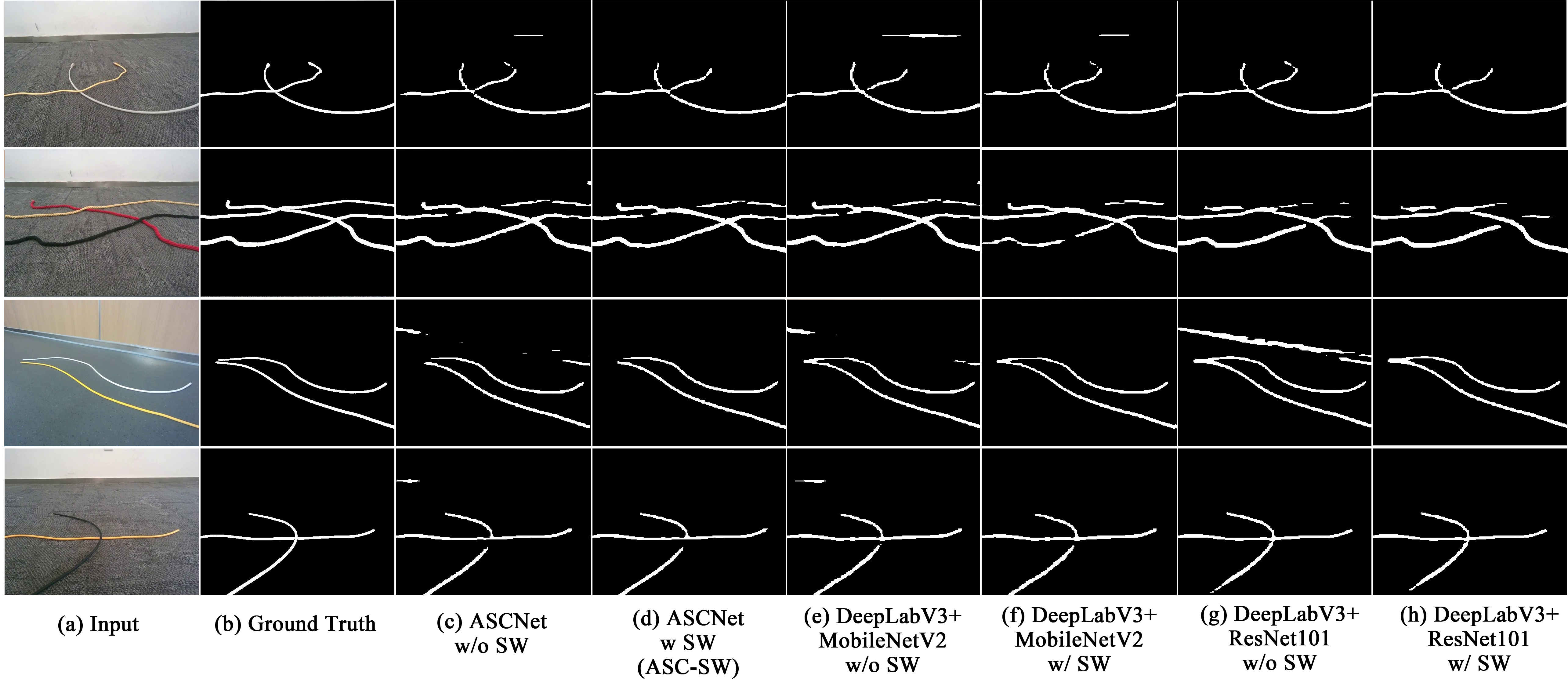}  
            \caption{Segmentation result with SW and without SW. (a): Input image; (b): Ground truth; (c): Result of ASCNet; (d): Result of ASC-SW; (e): Result of DeepLabV3+ (MobileNetV2); (f): Result of DeepLabV3+ (MobileNetV2) with SW;(g): Result of DeepLabV3+(ResNet101); (h): Result of DeepLabV3+(ResNet101) with SW.}
            \label{fig:after}
\end{figure*}
After post-processing, the depth camera extracts point clouds from the segmented mask.
\section{experiment}
\subsection{Experimental Setup}
Experiments are conducted on the Ubuntu 20.04 with Python 3.8 and a NVIDIA 3070 (8GB). The backbone uses pretrained weights. The training batch size sets 4, 200 epochs, and Adam optimizer with learning rate 1e-5. 
Based on parameter tuning experiments as shown in Fig.\ref{fig:para}.  We ultimately selected: sliding window size is 55, tolerant is 10 and dilation kernel is (1,1). SW is a framework-level post-processing module and is therefore not applied to other standalone segmentation models in the comparisons.
\subsection{Dataset}
\subsubsection{Training dataset}
\label{sec:dataset}
The training dataset extends the public set\cite{zanella2021auto} with 1,850 additional 1280×720 synthetic images, enriching samples of black and white DLOs across 10 backgrounds, all captured from a top-down viewpoint. 
\begin{figure*}[!htb]
            \centering
            \includegraphics[width=\textwidth]{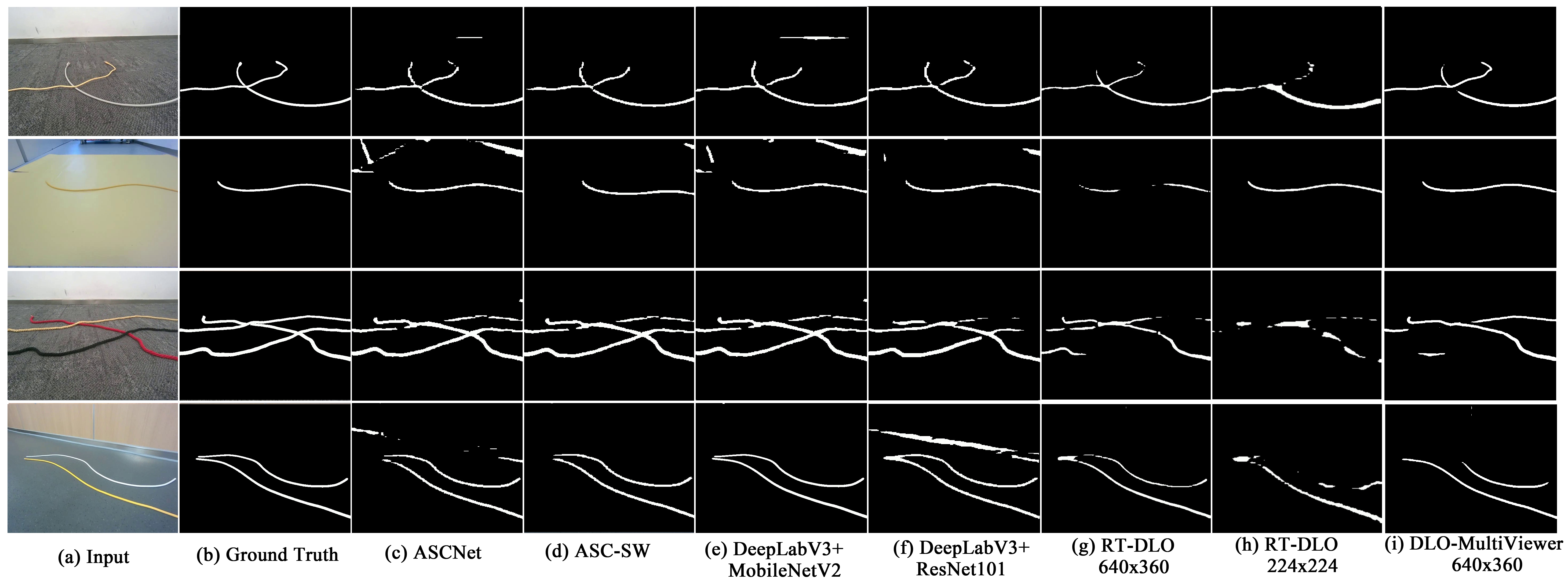}  
            \caption{Segmentation result of models. (a): Input image; (b): Ground truth; (c): Result of ASCNet; (d): Result of ASC-SW; (e): Result of DeepLabV3P+(MobileNetV2); (f): Result of DeepLabV3P+(ResNet101); (g): Result of RT-DLO with 640x360 input; (h): Result of RT-DLO with 224x224; (g): Result of DLO-MultiViewer with 640x360 input.}
            \label{fig:befor}
\end{figure*}
\subsubsection{Testing dataset}
The test set comprises 801 real-world 640×480 images from three indoor scenarios (mobile robot perspective (oblique view)), containing black, white, red and yellow DLOs (We collected four types of lines are shown in Fig.\ref{fig:lineappearance}a) on differently colored floors (black, white, and yellow) to assess robustness against noise. The images were collected by a mobile robot which is shown in Fig.\ref{fig:pipline} (lower left corner) using an Intel RealSense D455 and manually annotated. The test images are extracted from videos and evaluated in their original temporal order. The Fig.\ref{fig:lineappearance}b shows a significant shift in feature distribution between training and testing domains, confirming a non-trivial domain adaptation scenario.
\subsection{Comparison}
\begin{table*}[!htb]
    \centering
    \begin{threeparttable}
    \caption{The comparison of segmentation performance and efficiency}
     
    \begin{tabular}{l|cccccccc}
    \hline
        \textbf{Model} & \textbf{SW} & \textbf{Parameters} & \textbf{Backbone} & \textbf{Resolution} & \textbf{FLOPS (G)} & \textbf{Latency (ms)} & \textbf{Inference FPS} & \textbf{Mean IoU} \\ \hline
        DeepLabV3+\cite{chen2018encoder} & W/O & 59.9M & ResNet101 & 224x224 & 14.53  & 6.4  & 155  & 0.5861\\ 
        DeepLabV3+\cite{chen2018encoder} & W/ & 59.9M & ResNet101 & 224x224 & 14.56  & 6.8  & 147  & 0.6055\\ 
        RT-DLO\cite{caporali2023rt} & W/O & 46M & ResNet101 & 224x224 & 10.74  & 7.2  & 168  & 0.1817\\ 
        RT-DLO\cite{caporali2023rt} & W/O & 46M & ResNet101 & 640x360 & 50.18  & 19.4  & 50  & 0.6322\\ 
        DeepLabV3+\cite{chen2018encoder} & W/O & 6.8M & MobileNetV2 & 224x224 & 6.34  & 3.7  & 268  & 0.5953\\ 
        DeepLabV3+\cite{chen2018encoder} & W/ & 6.8M & MobileNetV2 & 224x224 & 6.38  & 3.8  & 266  & 0.6165
\\ 
        DLOPercevier\cite{caporali2024dlo} & W/O & 46M+66M\tnote{*} & Multimodel & 640x360 & 70.22  & 31.8  & 31  & 0.1254  \\ 
        Grounded-SAM\cite{ren2024grounded} & W/O & 649M+172M\tnote{*} & Multimodel & 1920x1080 & - & - & - & 0.4855  \\ 
        Grounded-SAM\cite{ren2024grounded} & W/O & 649M+172M & Multimodel & 224x224 & - & - & - & 0.2224  \\ 
        Grounded-FastSAM\cite{zhao2023fast} & W/O & 72M+172M\tnote{*}  & Multimodel & 224x192 & 17.35  & 9.1  & 109  & 0.0281  \\ 
        Grounded-FastSAM\cite{zhao2023fast} & W/O & 72M+172M & Multimodel & 1024x768 & 123.96  & 34.4  & 29  & 0.0271  \\ 
        DLO-MultiViewer\cite{caporali2025robotic} & W/O & 26M & ResNet50 & 640x360 & 32.71  & 13.3  & 75  & 0.6262\\ 
        DLO-MultiViewer\cite{caporali2025robotic} & W/O & 26M & ResNet50 & 224x224 & 7.02  & 3.6  & 275  & 0.4259  \\ \hline
        ASC-SW & W/ & 6.9M & MobileNetV2 & 224x224 & 6.39  & 4.1  & 245  & \textbf{0.7405}  \\ 
        ASCNet & W/O & 6.9M & MobileNetV2 & 224x224 & 6.36  & 3.8  & 261  & 0.6220 \\ \hline
    \end{tabular}
    \begin{tablenotes}
        \footnotesize
        \item[*] ResNet101: 46M, BERT: 66M, SAM: 649M, Grounding DINO: 172M, FastSAM: 72M 
    \end{tablenotes}

    \end{threeparttable}
    \label{tab:segmentation}
\end{table*}
We conducted a comprehensive evaluation of ASC-SW on the self-collected real-world dataset which in \cref{sec:dataset}. First, we compared ASC-SW with the RT-DLO model. Since ASC-SW operates with an input resolution of 224×224, RT-DLO was evaluated at two input resolutions, 640×360 (original resolution) and 224×224, to ensure a fair comparison.
In addition, we evaluated DeepLabV3+\cite{chen2018encoder} with two backbone networks (MobileNetV2 and ResNet-101), and models were trained with the same dataset as ASC-SW. The SW module was further integrated into these models to assess its impact on segmentation performance.
We compared ASC-SW with the 2D DLO segmentation model proposed in \cite{zhaole2023robust}, using input resolutions of 1920×1080 (original resolution) and 224×224. For Grounded-SAM\cite{ren2024grounded}, we replaced the original SAM model with the more lightweight FastSAM. Due to the architectural constraints of FastSAM\cite{zhao2023fast}, which only supports aspect-ratio-preserving input scaling, we evaluated it at 1024×768 (original resolution) and 224×192.
Furthermore, we selected the 2D DLO segmentation model from DLO-MultiViewer\cite{caporali2025robotic} and evaluated it at 640×360 (original resolution) and 224×224. We also included DLOPerceiver\cite{caporali2024dlo} in our comparison, which was evaluated using its original input resolution of 640×360.
Finally, to analyze the contribution of the Sliding Window module, we evaluated ASCNet without the SW module for comparison.
\subsubsection{Lightweighting comparison}
In Table \ref{tab:segmentation}, we evaluated all models on an RTX 3070 (except Grounded-SAM, which cannot run on the RTX 3070). The detailed configuration of the test platform is as follows: RTX 3070 GPU, Intel i9-13900 CPU, 32 GB RAM. In Table \ref{tab:edge}, we evaluated ASC-SW and DeepLabV3+ (MobileNetV2) the only two models deployable on edge devices on two edge platforms: Jetson TX1 and Jetson Orin Nano.
From Table \ref{tab:segmentation}, it can be observed that introducing the SW module results in a slight increase in inference latency and a minor decrease in frame rate, along with a small increase in computational cost. Moreover, reducing the input resolution significantly improves inference speed and reduces latency, indicating that low-resolution inputs are more suitable for deployment on edge devices. DeepLabV3+ (MobileNetV2) (with or without SW) is the most lightweight model, achieving the lowest inference latency and the highest frame rate, while ASCNet (with or without SW) performs only slightly lower, with a difference of less than 1\%.
In Table \ref{tab:edge}, DeepLabV3+ (MobileNetV2) remains the most lightweight model, and the performance gap between it and ASCNet is still less than 1\%. Compared with the original DeepLabV3+ (MobileNetV2), ASCNet achieves improved segmentation performance with almost no additional computational cost.
\subsubsection{Comparison of segmentation results}
  \begin{table*}[!htb]
    \centering
    \caption{The comparison of efficiency on edge device}
    \begin{tabular}{l|cccccc}
    \hline
        \textbf{Model} & \textbf{Backbone} & \textbf{SW} & \textbf{Resolution} & \textbf{Latency (ms)} & \textbf{Inference FPS} & \textbf{Device} \\ \hline
        DeepLabV3+ & MobileNetV2 & W/ & 224x224 & 150.62 & 6.64 & Jetson TX1 \\ 
        DeepLabV3+ & MobileNetV2 & W/ & 224x224 & 53.06 & 18.85 & Jetson Orin Nano \\ \hline
        ASCNet & MobileNetV2 & W/ & 224x224 & 154.96 & 6.45 & Jetson TX1 \\ 
        ASCNet & MobileNetV2 & W/ & 224x224 & 54.14 & 18.45 & Jetson Orin Nano \\ \hline
    \end{tabular}
    \label{tab:edge}
\end{table*}
As shown in the table \ref{tab:segmentation}, ASC-SW achieves the best overall segmentation performance, followed by RT-DLO at its original resolution, and remains the top-performing model even at 224×224, where other baselines degrade significantly.
The table\ref{tab:segmentation} also shows that when the camera viewpoint during inference differs substantially from the training data viewpoint, multi-model approaches perform noticeably worse than deep learning methods in segmentation. 
Furthermore, as shown in Fig.\ref{fig:befor}, RT-DLO performs poorly when the background color is similar to that of the DLOs. When the input resolution is reduced to 224×224 (the same as other self-trained models in this paper), the segmentation results degrade significantly. ASCNet, DeepLabV3+ (MobileNetV2), and DeepLabV3+ (ResNet101) all show varying degrees of false positives while successfully segmenting the DLOs, often incorrectly segmenting line-like areas in the scene. 

As illustrated in Fig.\ref{fig:befor}, it is more prone to incorrect segmentation due to interference from visually similar objects in the scene, resulting in lower generalization capability for lightweight networks such as ASCNet and DeepLabV3+ (MobileNetV2), both of which are trained with low-resolution input. This reduces the robustness and generalization of the models. In contrast, RT-DLO trained with high-resolution input (640×360) has higher latency but offers better generalization and produces more accurate segmentation masks with fewer incorrect segmentation. As shown in the Fig.\ref{fig:after}, the segmentation results of three models with and without the SW module are compared. The introduction of the SW module effectively suppresses most background noise, thereby improving the overall segmentation quality. ASCNet with SW post-processing achieves the best segmentation performance, with only minor false detections as shown in Fig.\ref{fig:after}. 
ASC-SW integrates the lightweight ASCNet with the SW algorithm, achieving accurate DLOs segmentation at low computational cost. ASCNet’s lightweight metrics are within 1\% of DeepLabV3+ (MobileNetV2), while delivering higher segmentation accuracy both before and after applying SW.
\subsubsection{Cross-View Performance Comparison}
 \begin{table}[!ht]
\centering
\caption{Comparison of models trained only on manipulator data and tested on mobile robot data}
    \label{tab:zhufan}
    \begin{tabular}{lccc}
    \hline
        \textbf{Model} &  \textbf{Training View} & \textbf{Testing View} & \textbf{Mean IoU} \\ \hline
        DeepLabV3+ & Manipulator & Mobile & 0.5953 \\  
        RT-DLO  & Manipulator & Mobile & 0.1817  \\ 
        ASCNet  & Manipulator & Mobile & 0.622 \\ 
        ASC-SW  & Manipulator & Mobile & \textbf{0.7405} \\ \hline
    \end{tabular}
\end{table}
As shown in Table \ref{tab:zhufan}, RT-DLO degrades significantly under viewpoint shifts, DeepLabV3+ produces false positives, while ASCNet preserves structural continuity and SW further improves robustness via temporal filtering. The method enhances DLO perception by aligning directional receptive fields with their geometry, integrating multi-scale atrous strip features, and enforcing temporal stability constraints.
\subsection{Ablation Study}
\subsubsection{Efficiency of CA}
\begin{table}[!ht]
    \centering
    \caption{Comparison between ASCSPP and other feature extraction modules.}
    \label{tab:ASCSPP}
    \begin{tabular}{ccccccc}
    \hline
        \textbf{ASPP} & \textbf{DAPP} & \textbf{SPASPP} & \textbf{ASCSPP} & \textbf{CA} & \textbf{Mean IoU}  \\ \hline
         $\checkmark$& ~ & ~ & ~ & $\checkmark$ & 0.5927  \\ 
        ~ &  $\checkmark$& ~ & ~ & $\checkmark$ & 0.5937  \\ 
        ~ & ~ &  $\checkmark$& ~ & $\checkmark$ & 0.5108  \\ 
        ~ & ~ & ~ &  $\checkmark$ & ~& 0.5847  \\ 
        ~ & ~ & ~ &  $\checkmark$ & $\checkmark$& 0.6220 \\
        \hline
    \end{tabular}
\end{table}
To verify the effectiveness of the modified CA module, we conducted a simple experiment. We tested the models with and without the CA module, and the results are shown in Table \ref{tab:ASCSPP}. The CA module brought improvements of 6\%.
\subsubsection{Efficiency of ASCSPP}
\begin{figure}[htbp]
            \centering
            \includegraphics[width=\columnwidth]{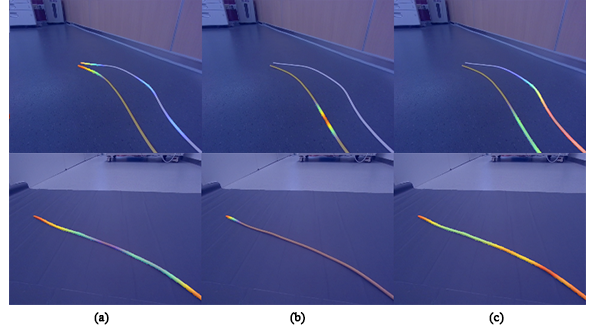} 
            \caption{Comparative Visualization of Intermediate Feature Responses in Different Fusion Modules: (a): Mask-conditioned Grad-CAM for DAPP; (b): Mask-conditioned Grad-CAM for SPASPP; (c): Mask-conditioned Grad-CAM for ASCSPP.} 
            \label{fig:ablation}
        \end{figure}
We trained ASCSPP, ASPP, DAPP, and SPASPP modules based on the ASCNet framework using the same training data and strategy, and compared them on the self-built real world dataset. The results verify that ASCSPP achieved the best segmentation performance (see Table \ref{tab:ASCSPP}). As shown in Fig.\ref{fig:ablation}. The atrous strip convolution \cref{sec:asconv} in ASCSPP makes it more sensitive to linear structures than DAPP and SPASPP
\section{Discussion}
\begin{figure}[htbp]
            \centering
            \includegraphics[width=\columnwidth]{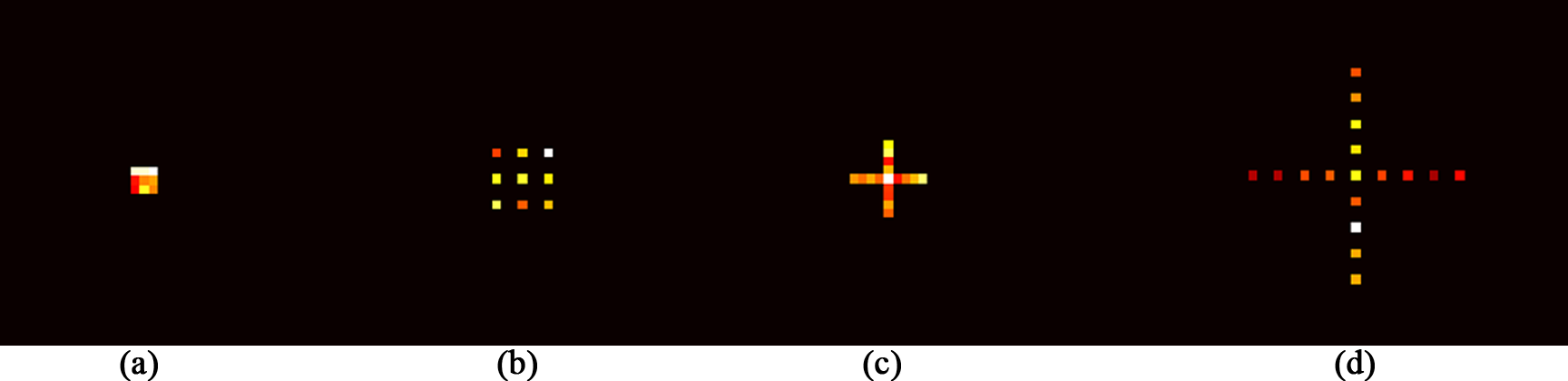} 
            \caption{(a–d): 3x3 ConV, 3x3 Dilated ConV, Strip ConV, ASConV} 
            \label{fig:erf}
        \end{figure}
DLOs exhibit high aspect ratios, strong directional continuity, thin spatial support, and sparse foreground occupancy. Traditional isotropic square convolution kernels (e.g., 3×3) are suboptimal for such elongated structures, as they tend to aggregate excessive background noise. In contrast, strip convolution employs 1D directional filters (horizontal and vertical) to form anisotropic receptive fields aligned with linear geometry, reducing parameter redundancy while enhancing directional sensitivity. 
Fig.\ref{fig:erf} visualizes that Atrous convolution enlarges the receptive field without reducing spatial resolution, which is essential for DLOs detection where long-range context is required to preserve continuity and distinguish thin wires from similar background structures. However, standard atrous convolution relies on isotropic square kernels, whose dilated receptive fields may introduce background interference. To address this, we propose ASConv, which expands the receptive field anisotropically while maintaining directional selectivity and reducing FLOPs compared to full 2D dilated kernels. As shown in Fig.\ref{fig:ablation}. This design is better aligned with the structural priors of the DLOs.
Compared with a 3×3 dilated convolution, strip convolution maintains a similar parameter order while reducing multiplications through kernel decomposition. Its effective receptive field is elongated along the principal direction, which suppresses background interference by enforcing directional filtering.
\section{Conclusion}
In this paper, we proposed an efficient segmentation model, ASCNet, based on the DeepLabV3+ architecture. ASCNet is designed to detect and segment DLOs, and we further proposed a post-processing algorithm (SW) to enhance its performance. When combined ASC-SW, the model can be trained using datasets collected from manipulator platforms with a fixed camera view, while still maintaining robustness under varying camera angles on mobile robots. ASC-SW strikes a balance between lightness and accuracy, achieving efficient segmentation of DLOs. It is much smaller in size than the baseline, and its segmentation performance is slightly better. The visual-assisted navigation framework can be deployed on edge computing devices and effectively avoids DLOs on the ground.
However, there is still room for improvement in ASCNet’s lightweight design. Although it can be deployed on edge devices, its inference speed has not yet reached real-time performance (30 FPS). Future work will continue to explore more efficient segmentation methods for DLOs.

\bibliographystyle{IEEEtran}
\bibliography{name}


\end{document}